# Improved Use of Continuous Attributes in C4.5

**J. R. Quinlan**                                                    QUINLAN@CS.SU.OZ.AU
*Basser Department of Computer Science*
*University of Sydney, Sydney Australia 2006*

## Abstract

A reported weakness of C4.5 in domains with continuous attributes is addressed by modifying the formation and evaluation of tests on continuous attributes. An MDL-inspired penalty is applied to such tests, eliminating some of them from consideration and altering the relative desirability of all tests. Empirical trials show that the modifications lead to smaller decision trees with higher predictive accuracies. Results also confirm that a new version of C4.5 incorporating these changes is superior to recent approaches that use global discretization and that construct small trees with multi-interval splits.

## 1. Introduction

Most empirical learning systems are given a set of pre-classified cases, each described by a vector of attribute values, and construct from them a mapping from attribute values to classes. The attributes used to describe cases can be grouped into *continuous* attributes, whose values are numeric, and *discrete* attributes with unordered nominal values. For example, the description of a person might include weight in kilograms, with a value such as 73.5, and color of eyes whose value is one of 'brown', 'blue', etc.

C4.5 (Quinlan, 1993) is one such system that learns decision-tree classifiers. Several authors have recently noted that C4.5's performance is weaker in domains with a preponderance of continuous attributes than for learning tasks that have mainly discrete attributes. For example, Auer, Holte, and Maass (1995) describe T2, a system that searches for good two-level decision trees, and comment:

> "The accuracy of T2's trees rivalled or surpassed C4.5's on 8 of the [15] datasets, including all but one of the datasets having only continuous attributes."

Discussing the effect of replacing continuous attributes by discrete attributes, each of whose values corresponds to an interval of the continuous attribute, Dougherty, Kohavi, and Sahami (1995) write:

> "C4.5's performance was significantly improved on two datasets ... using the entropy discretization method and did not significantly degrade on any dataset. ... We conjecture that the C4.5 induction algorithm is not taking full advantage of possible local discretization."

This paper explores a new version of C4.5 that changes the relative desirability of using continuous attributes. Section 2 sketches the current system, while the following section describes the modifications. Results from a comprehensive set of trials, reported in Section 4, show that the modifications lead to trees that are both smaller and more accurate. Section 5 compares the performance of the new version to results obtained with the two alternative methods of exploiting continuous attributes quoted above.

## 2. Constructing Decision Trees

C4.5 uses a divide-and-conquer approach to growing decision trees that was pioneered by Hunt and his co-workers (Hunt, Marin, & Stone, 1966). Only a brief description of the method is given here; see Quinlan (1993) for a more complete treatment.

The following algorithm generates a decision tree from a set $D$ of cases:

- If $D$ satisfies a *stopping criterion*, the tree for $D$ is a leaf associated with the most frequent class in $D$. One reason for stopping is that $D$ contains only cases of this class, but other criteria can also be formulated (see below).

- Some test $T$ with mutually exclusive outcomes $T_1, T_2, \ldots, T_k$ is used to partition $D$ into subsets $D_1, D_2, \ldots, D_k$, where $D_i$ contains those cases that have outcome $T_i$. The tree for $D$ has test $T$ as its root with one subtree for each outcome $T_i$ that is constructed by applying the same procedure recursively to the cases in $D_i$.

Provided that there are no cases with identical attribute values that belong to different classes, any test $T$ that produces a non-trivial partition of $D$ will eventually lead to single-class subsets as above. However, in the expectation that smaller trees are preferable (being easier to understand and often more accurate predictors), a family of possible tests is examined and one of them chosen to maximize the value of some *splitting criterion*. The default tests considered by C4.5 are:

- $A=?$ for a discrete attribute $A$, with one outcome for each value of $A$.

- $A \leq t$ for a continuous attribute $A$, with two outcomes, *true* and *false*. To find the threshold $t$ that maximizes the splitting criterion, the cases in $D$ are sorted on their values of attribute $A$ to give ordered distinct values $v_1, v_2, \ldots, v_N$. Every pair of adjacent values suggests a potential threshold $t = (v_i + v_{i+1})/2$ and a corresponding partition of $D$.[1] The threshold that yields the best value of the splitting criterion is then selected.

The default splitting criterion used by C4.5 is *gain ratio*, an information-based measure that takes into account different numbers (and different probabilities) of test outcomes. Let $C$ denote the number of classes and $p(D, j)$ the proportion of cases in $D$ that belong to the $j$th class. The residual uncertainty about the class to which a case in $D$ belongs can be expressed as

$$Info(D) = -\sum_{j=1}^{C} p(D, j) \times \log_2(p(D, j))$$

---

[1]. Fayyad and Irani (1992) prove that, for convex splitting criteria such as information gain, it is not necessary to examine all such thresholds. If all cases with value $v_i$ and with adjacent value $v_{i+1}$ belong to the same class, a threshold between them cannot lead to a partition that has the maximum value of the criterion.

and the corresponding information gained by a test $T$ with $k$ outcomes as

$$Gain(D, T) = Info(D) - \sum_{i=1}^{k} \frac{|D_i|}{|D|} \times Info(D_i) .$$

The information gained by a test is strongly affected by the number of outcomes and is maximal when there is one case in each subset $D_i$. On the other hand, the potential information obtained by partitioning a set of cases is based on knowing the subset $D_i$ into which a case falls; this *split information*

$$Split(D, T) = - \sum_{i=1}^{k} \frac{|D_i|}{|D|} \times \log_2 \left( \frac{|D_i|}{|D|} \right)$$

tends to increase with the number of outcomes of a test. The gain ratio criterion assesses the desirability of a test as the ratio of its information gain to its split information. The gain ratio of every possible test is determined and, among those with at least average gain, the split with maximum gain ratio is selected.

In some situations, every possible test splits $D$ into subsets that have the same class distribution. All tests then have zero gain, and C4.5 uses this as an additional stopping criterion.

The recursive partitioning strategy above results in trees that are consistent with the training data, if this is possible. In practical applications data are often noisy – attribute values are incorrectly recorded and cases are misclassified. Noise leads to overly complex trees that attempt to account for these anomalies. Most systems *prune* the initial tree, identifying subtrees that contribute little to predictive accuracy and replacing each by a leaf.

## 3. Modified Assessment of Continuous Attributes

We return now to the selection of a threshold for a continuous attribute $A$. If there are $N$ distinct values of $A$ in the set of cases $D$, there are $N - 1$ thresholds that could be used for a test on $A$. Each threshold gives unique subsets $D_1$ and $D_2$ and so the value of the splitting criterion is a function of the threshold. The ability to choose the threshold $t$ so as to maximize this value gives a continuous attribute $A$ an advantage over a discrete attribute (which has no similar parameter that adjusts the partition of $D$), and also over other continuous attributes that have fewer distinct values in $D$. That is, the choice of a test will be biased towards continuous attributes with numerous distinct values.

This paper proposes a correction for this bias that consists of two modifications to C4.5. The first of these, inspired by the Minimum Description Length principle (Rissanen, 1983), adjusts the apparent information gain from a test of a continuous attribute. Discussion of this change is prefaced by a brief introduction to MDL.

Following Quinlan and Rivest (1989), let a *sender* and a *receiver* both possess an ordered list of the cases in the training data showing each case's attribute values. The sender also knows the class to which each case belongs and must transmit this information to the receiver. He or she first encodes and sends a *theory* of how to classify the cases. Since this theory might be imperfect, the sender must also identify the *exceptions* to the theory that occur in the training cases and state how their classes predicted by the theory should be corrected. The total length of the transmission is thus the number of bits required to encode the theory (the *theory cost*) plus the bits needed to identify and correct the exceptions (the *exceptions cost*). The sender may have a choice among several alternative theories, some being simple but leaving many errors to be corrected while others are more elaborate but more accurate. The MDL principle may then be stated as: Choose the theory that minimizes the sum of the theory and exceptions costs.

MDL thus provides a framework for trading off the complexity of a theory against its accuracy on the training data $D$. The exceptions cost associated with a set of cases $D$ is asymptotically equivalent to $|D| \times Info(D)$, so that $|D| \times Gain(D, T)$ measures the reduction in exceptions cost when $D$ is partitioned by a test $T$. Partitioning $D$ in this way, however, requires transmission of a more complex theory that includes the definition of $T$. Whereas a test $A=?$ on a discrete attribute $A$ can be specified by nominating the attribute involved, a test $A \leq t$ must also include the threshold $t$; if there are $N-1$ possible thresholds for $A$, this will take an additional $\log_2(N-1)$ bits.[2] The first modification is to "charge" this increased cost associated with a test on a continuous attribute to the apparent gain achieved by the test, so reducing the (per-case) information $Gain(D, T)$ by $\log_2(N-1)/|D|$.

A test on a continuous attribute with numerous distinct values will now be less likely to have the maximum value of the splitting criterion among the family of possible tests, and so is less likely to be selected. Further, if all thresholds $t$ on a continuous attribute $A$ have an adjusted gain that is less than zero, attribute $A$ is effectively ruled out. The consequences of this first change are thus a re-ranking of potential tests and the possible exclusion of some of them.

The second change is much more straightforward. Recall that the gain ratio criterion divides the apparent gain by the information available from a split. This latter varies as a function of the threshold $t$ and is is maximal when there are as many cases above $t$ as below. If the gain ratio criterion is used to select $t$, the effect of the penalty described above will also vary with $t$, having the least impact when $t$ divides the cases equally. This seems to be an unnecessary complication, so the threshold $t$ is chosen instead to maximize gain. Once the threshold is chosen, however, the final selection of the attribute to be used for the test is still made on the basis of the gain ratio criterion using the adjusted gain.

## 4. Empirical Evaluation

The effects of these changes were assessed empirically in a series of "before and after" experiments with a substantial number of learning tasks. Release 7 of C4.5 (abbreviated here as Rel 7) was modified as above to produce a new version (Rel 8). Both systems were applied to twenty databases from the UCI Repository that involve continuous attributes, either alone or in combination with discrete attributes. A summary of the characteristics of these data sets appears in Appendix A. In all the following experiments, both versions of C4.5 were run with the same default settings for all parameters; no attempt was made to tune either system for these tasks.

---

2. Even with a convex splitting criterion that satisfies the requirements of Fayyad and Irani (1992), we cannot use the number $N'$ of potentially gain-maximizing thresholds instead of the greater number $N$ of possible thresholds. Since the receiver knows the cases' attribute values but not their classes, he or she cannot determine whether all the cases with two adjacent values of $A$ belong to the same class. The message must consequently identify the chosen threshold among all possible thresholds.

Table 1: Results using modified (*Rel 8*) and previous (*Rel 7*) C4.5.

| | Error Rate | | | | Tree Size | | |
|---|---|---|---|---|---|---|---|
| | Rel 8 | Rel 7 | w/d/l | ratio | Rel 8 | Rel 7 | ratio |
| anneal | 7.67 ±.12 | 7.49 ±.16 | 3/2/5 | 1.02 | 75.2 ±.7 | 70.1 ±1.1 | 1.07 |
| auto | 17.7 ±.5 | 23.8 ±.6 | 10/0/0 | .74 | 63.7 ±.4 | 62.9 ±.5 | 1.01 |
| breast-w | 5.26 ±.19 | 5.29 ±.09 | 5/1/4 | .99 | 25.0 ±.5 | 20.3 ±.5 | 1.23 |
| colic | 15.0 ±.2 | 15.1 ±.4 | 5/2/3 | .99 | 9.7 ±.2 | 20.0 ±.5 | .49 |
| credit-a | 14.7 ±.2 | 15.8 ±.3 | 7/1/2 | .93 | 33.2 ±1.1 | 57.3 ±1.2 | .58 |
| credit-g | 28.4 ±.3 | 28.9 ±.3 | 5/1/4 | .98 | 124 ±2 | 155 ±2 | .80 |
| diabetes | 25.4 ±.3 | 28.3 ±.3 | 10/0/0 | .90 | 44.0 ±1.6 | 128.2 ±1.8 | .34 |
| glass | 32.5 ±.8 | 32.1 ±.5 | 4/1/5 | 1.01 | 45.7 ±.4 | 51.3 ±.4 | .89 |
| heart-c | 23.0 ±.5 | 24.9 ±.4 | 8/0/2 | .92 | 39.9 ±.4 | 45.3 ±.3 | .88 |
| heart-h | 21.5 ±.2 | 21.6 ±.5 | 4/0/6 | 1.00 | 19.1 ±.6 | 29.7 ±1.2 | .64 |
| hepatitis | 20.4 ±.6 | 21.7 ±.8 | 6/1/3 | .94 | 17.8 ±.3 | 15.5 ±.4 | 1.15 |
| hypo | .48 ±.01 | .49 ±.02 | 6/3/1 | .97 | 27.5 ±.1 | 25.3 ±.1 | 1.09 |
| iris | 4.80 ±.17 | 4.87 ±.20 | 3/3/4 | .99 | 8.5 ±.0 | 9.3 ±.1 | .91 |
| labor | 19.1 ±1.0 | 16.7 ±.9 | 1/2/7 | 1.15 | 7.0 ±.3 | 7.3 ±.1 | .96 |
| letter | 12.0 ±.0 | 12.2 ±.0 | 10/0/0 | .98 | 2328 ±4 | 2370 ±4 | .98 |
| segment | 3.21 ±.08 | 3.77 ±.07 | 9/1/0 | .85 | 82.9 ±.5 | 83.5 ±.6 | .99 |
| sick | 1.34 ±.03 | 1.29 ±.03 | 2/1/7 | 1.04 | 50.8 ±.5 | 51.5 ±.5 | .99 |
| sonar | 25.6 ±.7 | 28.4 ±.6 | 8/0/2 | .90 | 28.4 ±.2 | 33.1 ±.5 | .86 |
| vehicle | 27.1 ±.4 | 29.1 ±.3 | 10/0/0 | .93 | 135 ±2 | 181 ±1 | .75 |
| waveform | 27.3 ±.3 | 28.1 ±.6 | 6/2/2 | .97 | 44.6 ±.4 | 49.2 ±.4 | .91 |
| **average** | | | | **.96** | | | **.88** |

## 4.1 Initial experiments

Table 1 displays the results of the first trials, consisting of ten complete ten-fold cross-validations[3] with each task. The figure shown for each system is the mean of the ten cross-validation results where the error rates and tree sizes refer to C4.5's pruned trees; the standard error of this mean appears in small font. The column headed 'w/d/l' shows the number of complete cross-validations in which Rel 8 gives a lower error rate, the same error rate, or a higher error rate than Rel 7. The figures under 'ratio' present results for Rel 8 divided by the corresponding figure for Rel 7.

As the overall averages at the foot of the table indicate, the trees produced by Rel 8 in these trials are 4% more accurate and 12% smaller than those generated by Rel 7. Rel 8 is less accurate that Rel 7 on only four of the twenty tasks; for the smallest data set (labor, with 57 cases), however, the trees produced by Rel 8 are substantially less accurate. The pruned trees generated by Rel 8 for some tasks are a great deal smaller than their Rel 7 counterparts – diabetes is a particularly notable example.

---

3. A ten-fold cross-validation is performed by dividing the data into ten blocks of cases that have similar size and class distribution. For each block in turn, a decision tree is constructed from the remaining nine blocks and tested on the unseen cases in the hold-out block.

I do not recommend the use of the unpruned trees constructed initially by C4.5 but, for the sake of completeness, the corresponding figures for the unpruned trees were also determined. The average ratio of the error rate of Rel 8 to that of Rel 7 is 0.95, while the ratio of tree size is 0.94. For the unpruned trees, then, the modifications incorporated in Rel 8 lead to a 5% reduction in error and a 6% reduction in the size.

## 4.2 Adding irrelevant attributes

In practical applications, it is unlikely that an analyst would knowingly add irrelevant attributes to the data! However, even an attribute that is relevant for some parts of the tree might be quite irrelevant for others. The bias towards continuous attributes inherent in Rel 7 implies that the system should occasionally select a test on an irrelevant continuous attribute in preference to tests on relevant discrete attributes.

To explore this potential deficiency, the twenty data sets were modified by adding irrelevant attributes. Ten of these were continuous attributes, each having uniformly distributed random values $x$, $0 \leq x < 1$. (Since only the order of values of a continuous attribute is important, the distribution of these values does not matter – use of another distribution such as the Gaussian $N(0, 1)$ should produce comparable results.) As Kohavi (personal communication, 1995) points out, it is unfair to compare Rel 8 to Rel 7 on data sets to which only irrelevant continuous-valued attributes have been added, since the modifications incorporated in Rel 8 make it less likely to choose tests involving any continuous attributes. To circumvent this problem, a further ten discrete attributes with ten equiprobable values were added, giving twenty irrelevant attributes in all. The experiments above were repeated on the enlarged data sets, with the results shown in Table 2.

These results highlight the effects of the modifications implemented in Rel 8. Addition of irrelevant attributes increases the error of the Rel 7 trees by an average of 12%, but has a much smaller impact on those produced by Rel 8. The head-to-head comparison on the altered data sets, presented in the table, shows that the pruned trees found by Rel 8 have 10% lower error on average, and are also a great deal smaller. Any split on a random continuous attribute is unlikely to generate sufficient gain to "pay for" the threshold, so such tests will tend to be filtered out by Rel 8 but not by Rel 7. Consequently, Rel 7 is more prone to split the data (uselessly) on a random attribute, leading to larger trees and higher error rates.

## 4.3 Ablation experiments

The effects of the modifications implemented in Rel 8 can be factored into choosing (slightly) different thresholds using gain rather than gain ratio, excluding attributes for which no threshold gives sufficient gain to offset the penalty, and re-ranking potential tests by penalizing those that involve continuous attributes. To ascertain the contributions of each, two intermediate versions of C4.5 were constructed:

Table 2: Results after addition of irrelevant attributes.

| | Error Rate | | | | Tree Size | | |
|---|---|---|---|---|---|---|---|
| | Rel 8 | Rel 7 | w/d/l | ratio | Rel 8 | Rel 7 | ratio |
| anneal+ | 7.72 ±.23 | 8.13 ±.18 | 9/0/1 | .95 | 74.3 ±1.1 | 84.0 ±1.6 | .88 |
| auto+ | 18.7 ±.5 | 26.0 ±.7 | 10/0/0 | .72 | 63.4 ±.7 | 62.3 ±.6 | 1.02 |
| breast-w+ | 5.69 ±.11 | 6.17 ±.13 | 8/0/2 | .92 | 16.8 ±.4 | 25.0 ±.4 | .67 |
| colic+ | 15.1 ±.2 | 20.1 ±.3 | 10/0/0 | .75 | 8.9 ±.2 | 39.9 ±1.1 | .22 |
| credit-a+ | 13.6 ±.3 | 16.4 ±.3 | 10/0/0 | .83 | 34.7 ±.7 | 58.4 ±.9 | .60 |
| credit-g+ | 28.5 ±.3 | 32.4 ±.4 | 10/0/0 | .88 | 111 ±3 | 174 ±2 | .64 |
| diabetes+ | 26.9 ±.3 | 30.3 ±.5 | 10/0/0 | .89 | 43.6 ±2.1 | 115.5 ±1.8 | .38 |
| glass+ | 37.0 ±.5 | 35.9 ±.8 | 3/0/7 | 1.03 | 31.0 ±.6 | 46.2 ±.9 | .67 |
| heart-c+ | 22.6 ±.7 | 30.3 ±.4 | 10/0/0 | .75 | 24.9 ±.8 | 52.0 ±.5 | .48 |
| heart-h+ | 20.3 ±.4 | 24.9 ±.5 | 10/0/0 | .82 | 19.9 ±.5 | 32.0 ±.9 | .62 |
| hepatitis+ | 19.1 ±.6 | 23.9 ±.7 | 10/0/0 | .80 | 5.6 ±.4 | 20.4 ±.6 | .28 |
| hypo+ | .47 ±.02 | .49 ±.02 | 7/2/1 | .96 | 27.8 ±.2 | 25.9 ±.2 | 1.08 |
| iris+ | 5.67 ±.15 | 5.73 ±.41 | 4/0/6 | .99 | 7.6 ±.1 | 9.7 ±.1 | .79 |
| labor+ | 19.1 ±.8 | 24.6 ±.7 | 9/1/0 | .78 | 6.8 ±.2 | 10.6 ±.1 | .64 |
| letter+ | 12.7 ±.1 | 13.3 ±.1 | 10/0/0 | .95 | 2300 ±3 | 2372 ±6 | .97 |
| segment+ | 3.91 ±.09 | 3.85 ±.05 | 4/0/6 | 1.01 | 69.2 ±.6 | 88.2 ±.6 | .78 |
| sick+ | 1.61 ±.05 | 1.57 ±.05 | 4/1/5 | 1.02 | 37.1 ±.8 | 54.8 ±.7 | .68 |
| sonar+ | 25.5 ±.8 | 29.3 ±.6 | 9/1/0 | .87 | 20.1 ±.4 | 34.0 ±.5 | .59 |
| vehicle+ | 28.7 ±.3 | 28.8 ±.2 | 5/3/2 | .99 | 109 ±1 | 162 ±1 | .67 |
| waveform+ | 30.1 ±.7 | 28.0 ±.6 | 4/0/6 | 1.08 | 27.9 ±1.0 | 48.9 ±.5 | .57 |
| **average** | | | | **.90** | | | **.66** |

- 7G differs from Rel 7 only in that the threshold $t$ is chosen to maximize information gain rather than gain ratio;

- 7GS also chooses thresholds on gain; if the gain of the best threshold is less than the penalty $\log_2(N-1)/|D|$, however, the test is excluded.

The only difference between 7GS and Rel 8 is the latter's application of the penalty when determining the relative desirability of possible tests.

The trials were repeated using the same cross-validation blocks as before for these intermediate versions. Average error rates, tree sizes, and ratios (again computed with respect to Rel 7) are presented in Table 3 and summarized graphically in Figure 1.

Selection of thresholds by gain rather than gain ratio (7G) has very little impact – the average error rate and tree size ratios with respect to Rel 7 are both very close to one. There are non-trivial changes for some tasks, however; for instance, the error rate on the segment data is considerably lower and the trees found for the breast-w task are noticeably larger.

Table 3: Results for intermediate systems 7G and 7GS.

| | Error Rate | | | | Tree Size | | | |
|---|---|---|---|---|---|---|---|---|
| | 7G | ratio | 7GS | ratio | 7G | ratio | 7GS | ratio |
| anneal | 7.73 ±.16 | 1.03 | 7.62 ±.16 | 1.02 | 73.9 ±.7 | 1.05 | 73.4 ±.7 | 1.05 |
| auto | 23.0 ±.9 | .97 | 22.7 ±.8 | .95 | 59.5 ±.9 | .95 | 59.0 ±.9 | .94 |
| breast-w | 5.21 ±.23 | .98 | 5.32 ±.17 | 1.01 | 24.4 ±.3 | 1.21 | 24.1 ±.5 | 1.19 |
| colic | 15.0 ±.4 | .99 | 15.0 ±.3 | .99 | 20.2 ±.5 | 1.01 | 17.9 ±.6 | .90 |
| credit-a | 14.7 ±.3 | .93 | 14.1 ±.2 | .89 | 50.1 ±1.0 | .88 | 38.3 ±1.0 | .67 |
| credit-g | 29.7 ±.3 | 1.03 | 29.1 ±.2 | 1.01 | 148 ±2 | .96 | 138 ±2 | .89 |
| diabetes | 27.1 ±.4 | .96 | 25.2 ±.3 | .89 | 127 ±2 | .99 | 45.4 ±2.0 | .35 |
| glass | 30.9 ±.5 | .96 | 31.3 ±.7 | .97 | 50.3 ±.3 | .98 | 46.2 ±.6 | .90 |
| heart-c | 25.0 ±.4 | 1.01 | 23.8 ±.5 | .96 | 44.5 ±.8 | .98 | 42.2 ±.9 | .93 |
| heart-h | 22.2 ±.4 | 1.03 | 20.9 ±.3 | .97 | 30.2 ±1.1 | 1.02 | 19.3 ±.7 | .65 |
| hepatitis | 22.0 ±.8 | 1.01 | 21.4 ±.6 | .98 | 17.3 ±.4 | 1.12 | 15.2 ±.4 | .98 |
| hypo | .49 ±.02 | 1.00 | .50 ±.01 | 1.02 | 25.7 ±.2 | 1.02 | 27.0 ±.1 | 1.07 |
| iris | 4.93 ±.23 | 1.01 | 4.80 ±.17 | .99 | 8.5 ±.1 | .92 | 8.5 ±.0 | .92 |
| labor | 18.8 ±1.1 | 1.13 | 19.5 ±1.0 | 1.17 | 7.8 ±.1 | 1.06 | 7.6 ±.1 | 1.04 |
| letter | 12.3 ±.0 | 1.00 | 12.2 ±.0 | 1.00 | 2327 ±3 | .98 | 2330 ±3 | .98 |
| segment | 3.39 ±.08 | .90 | 3.36 ±.07 | .89 | 83.7 ±.3 | 1.00 | 82.8 ±.3 | .99 |
| sick | 1.34 ±.02 | 1.04 | 1.34 ±.02 | 1.04 | 50.4 ±.3 | .98 | 51.4 ±.3 | 1.00 |
| sonar | 28.8 ±1.2 | 1.02 | 26.6 ±1.1 | .94 | 32.6 ±.4 | .98 | 28.5 ±.3 | .86 |
| vehicle | 28.1 ±.2 | .97 | 27.6 ±.3 | .95 | 178 ±1 | .99 | 145 ±2 | .80 |
| waveform | 28.1 ±.8 | 1.00 | 27.3 ±.6 | .97 | 46.8 ±.4 | .95 | 44.4 ±.6 | .90 |
| **average** | | **1.00** | | **.98** | | **1.00** | | **.90** |

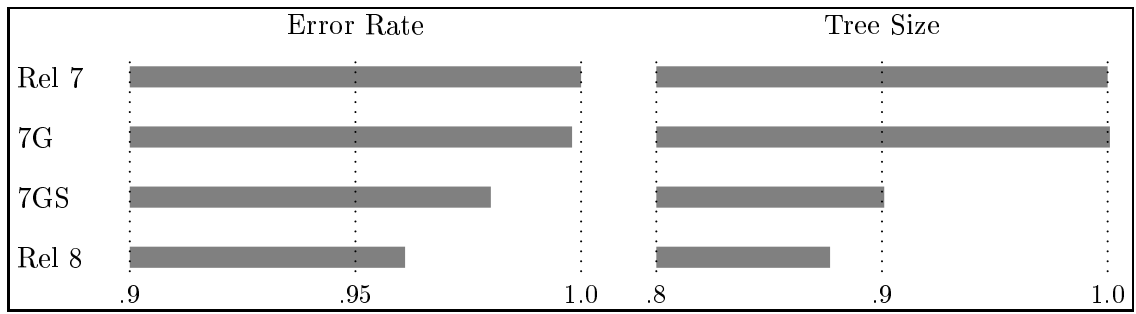

Figure 1: Summary of ablation results.

Use of the penalty to filter tests on continuous attributes (7GS) produces more notice-able differences. Ruling out some tests on continuous attributes accounts for most of the reduction in tree size observed with Rel 8. In some cases, the trees are markedly smaller – for the diabetes data, the 7GS trees are on average only one-third of the size of those produced by Rel 7. This change also accounts for about half of Rel 8's improvement in error rate, the diabetes data again providing the greatest change from 7G.

Finally, the use by Rel 8 of the penalty to re-rank the attributes yields a further im-provement in error rate and a small decrease in average tree size. This re-ranking may be beneficial even when all attributes are continuous: the average error rate of Rel 8 is about 1% lower than that of 7GS on the nine tasks of this kind, in only two of which does 7GS give a lower error rate than Rel 8.

## 5. Related Research

This section examines the two alternative methods for utilizing continuous attributes that were mentioned in the introduction, and compares them empirically with C4.5 Rel 8.

### 5.1 Global discretization

Dougherty et al. (1995) consider various ways of converting a continuous attribute to a discrete one by dividing its values into intervals, each of which becomes a separate value for the replacement discrete attribute. The method found to give the best results, entropy discretization, was first investigated by Catlett (1991) as a means of reducing the time required to construct a tree. Fayyad and Irani (1993) subsequently introduced a clever refinement that led to the final form used by Dougherty et al. and in the experiments reported here.

To find the set of intervals, the training cases are first sorted on the value of the continu-ous attribute in question. The procedure outlined in Section 2 is used to find the threshold $t$ that maximizes information gain. The same process is repeated for the corresponding subsets of cases with attribute values below and above $t$. (Since the cases are not reordered, they need not be re-sorted, and this is the source of the reduced learning times.) If $w$ thresholds are found, the continuous attribute is mapped to a discrete attribute with $w+1$ values, one for each interval.

Some stopping criterion is required to prevent this process from resulting in a very large number of intervals (which could become as numerous as the training cases if all values of the attribute are distinct). Catlett uses a four-pronged heuristic criterion, but Fayyad and Irani developed an elegant test based on the MDL principle (Section 3). They view a discretization rule as a classifying theory that uses a single attribute and that associates a class with each interval. Introduction of an additional threshold, increasing the complexity of the discretization rule, is allowed only if the greater theory coding cost is more than offset by the consequent reduction in the exceptions cost. This scheme generally leads to few thresholds in regions where the the cases' classes do not vary much and to finer divisions when required.

Similar experiments to those described by Dougherty et al. were carried out on the learning tasks of Section 4. In each trial, the training data are used to find discretization rules to convert every continuous attribute to a discrete attribute. C4.5[4] is invoked to find a tree that is evaluated on the test data, using the same discretization intervals found from the training data. As before, each data set is subjected to ten cross-validations using the same blocks of cases as previously.

Table 4: Comparison with C4.5 using global discretization (*Discr*).

| | Error Rate | | | | Tree Size | | |
|---|---|---|---|---|---|---|---|
| | Rel 8 | Discr | w/d/l | ratio | Rel 8 | Discr | ratio |
| anneal | 7.67 ±.12 | 9.48 ±.14 | 10/0/0 | .81 | 75.2 ±.7 | 68.1 ±.5 | 1.11 |
| auto | 17.7 ±.5 | 23.8 ±.6 | 9/1/0 | .74 | 63.7 ±.4 | 94.8 ±1.8 | .67 |
| breast-w | 5.26 ±.19 | 5.38 ±.15 | 6/0/4 | .98 | 25.0 ±.5 | 19.9 ±.5 | 1.25 |
| colic | 15.0 ±.2 | 15.1 ±.1 | 6/2/2 | .99 | 9.7 ±.2 | 7.8 ±.2 | 1.23 |
| credit-a | 14.7 ±.2 | 14.0 ±.1 | 0/1/9 | 1.05 | 33.2 ±1.1 | 22.3 ±.6 | 1.49 |
| credit-g | 28.4 ±.3 | 28.1 ±.4 | 5/1/4 | 1.01 | 124 ±2 | 82 ±1 | 1.50 |
| diabetes | 25.4 ±.3 | 25.5 ±.3 | 5/0/5 | .99 | 44.0 ±1.6 | 19.6 ±.7 | 2.25 |
| glass | 32.5 ±.8 | 28.4 ±.3 | 1/0/9 | 1.14 | 45.7 ±.4 | 35.8 ±.3 | 1.28 |
| heart-c | 23.0 ±.5 | 21.7 ±.6 | 2/1/7 | 1.06 | 39.9 ±.4 | 25.9 ±.4 | 1.54 |
| heart-h | 21.5 ±.2 | 20.8 ±.4 | 3/0/7 | 1.04 | 19.1 ±.6 | 9.7 ±.6 | 1.97 |
| hepatitis | 20.4 ±.6 | 19.6 ±.8 | 3/1/6 | 1.04 | 17.8 ±.3 | 11.5 ±.5 | 1.55 |
| hypo | .48 ±.01 | .72 ±.03 | 10/0/0 | .67 | 27.5 ±.1 | 45.1 ±.3 | .61 |
| iris | 4.80 ±.17 | 5.47 ±.29 | 6/3/1 | .88 | 8.5 ±.0 | 6.2 ±.1 | 1.36 |
| labor | 19.1 ±1.0 | 20.0 ±.9 | 6/0/4 | .96 | 7.0 ±.3 | 5.2 ±.1 | 1.34 |
| letter | 12.0 ±.0 | 21.1 ±.0 | 10/0/0 | .57 | 2328 ±4 | 9600 ±12 | .24 |
| segment | 3.21 ±.08 | 5.65 ±.10 | 10/0/0 | .57 | 82.9 ±.5 | 296.4 ±2.6 | .28 |
| sick | 1.34 ±.03 | 2.14 ±.03 | 10/0/0 | .63 | 50.8 ±.5 | 32.8 ±.4 | 1.55 |
| sonar | 25.6 ±.7 | 24.6 ±.7 | 3/1/6 | 1.04 | 28.4 ±.2 | 28.6 ±.5 | .99 |
| vehicle | 27.1 ±.4 | 31.5 ±.5 | 10/0/0 | .86 | 135 ±2 | 175 ±2 | .78 |
| waveform | 27.3 ±.3 | 26.5 ±.6 | 4/0/6 | 1.03 | 44.6 ±.4 | 42.2 ±.8 | 1.06 |
| **average** | | | | **.90** | | | **1.20** |

Results of these trials, summarized in Table 4, show that the comments of Dougherty et al. quoted in the introduction do not apply to Rel 8. Discretization leads to improved accuracy on eight of the tasks and to a degradation on 12 of them. Most of the improvements are modest, however, while several tasks exhibit a marked increase in error; the average value of the error ratio indicates a strong advantage for the local threshold selection employed in Rel 8 over the global thresholding used by discretization.

Kohavi (personal communication, 1996) suggests that there might be a "middle ground" in which thresholds are determined locally until the subsets of cases are relatively small, at which point subsequent possible thresholds would be found using the discretization strategy above. Evidence in support of this idea is provided by Figure 2 where, for each task, the error ratio that appears in Table 4 is plotted against the size of the data set (on a logarithmic scale). The clear trend shows that global discretization degrades performance more as data sets become larger, but can be beneficial for tasks with fewer cases.

---

4. Since there are no continuous attributes, Rel 7 and Rel 8 give identical results on these discretized tasks.

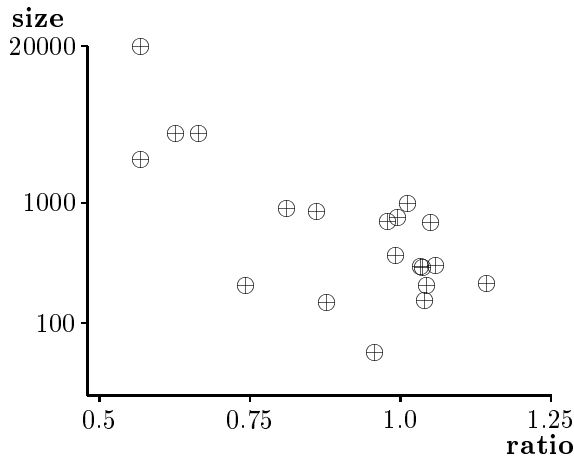

Figure 2: Effect of discretization vs data set size.

## 5.2 Multi-threshold splits

In contrast, T2 (Auer et al., 1995) determines thresholds locally but allows the values of a continuous attribute to be partitioned into multiple intervals. These intervals are not found heuristically by a recursive application of binary splitting, as above. Instead, a more thorough exploration is carried out to find the set of up to $m$ intervals that minimizes error on the training set. (The default value of $m$ is $C+1$ where there are $C$ classes in the data.) Search for these intervals is expensive, so T2 restricts decision trees to two levels of tests (in the spirit of one-level decision "stumps" described by Holte, 1993) where only the second level employs non-binary splits of continuous attributes. Within this restricted theory language, however, T2 is guaranteed to find a tree that misclassifies as few of the training cases as possible.

Even so, the computational cost of T2 using the default value of $m$ is proportional to $C^4 \cdot (C+1)^2 \cdot a^2$, where $a$ is the number of attributes (Auer, personal communication, 1996). For example, the time required to process the small auto data set with six classes and 25 attributes is four orders of magnitude greater than that needed by C4.5. This effectively rules out trials of T2 on some of the learning tasks used above, specifically those with more than four classes. For the remaining 14 tasks, experiments following the same pattern as before and using the same cross-validation blocks were carried out and are reported in Table 5. T2 produces trees with error rates much lower than those generated by Rel 8 on two tasks, slightly lower on two more, and higher on the remaining ten. As reflected in the average error ratio, the trials still favor C4.5 Rel 8 overall. (Had it been possible to run the tasks with larger numbers of classes, T2's restricted theory language would perhaps have caused an even more noticeable increase in error.)

Table 5: Comparison with T2.

| | Error Rate | | | | Tree Size | | |
|---|---|---|---|---|---|---|---|
| | Rel 8 | T2 | w/d/l | ratio | Rel 8 | T2 | ratio |
| breast-w | 5.26 ±.19 | 4.06 ±.09 | 0/0/10 | 1.30 | 25.0 ±.5 | 10.0 ±.0 | 2.50 |
| colic | 15.0 ±.2 | 16.2 ±.2 | 10/0/0 | .92 | 9.7 ±.2 | 15.5 ±.2 | .63 |
| credit-a | 14.7 ±.2 | 16.6 ±.2 | 10/0/0 | .89 | 33.2 ±1.1 | 46.1 ±.4 | .72 |
| credit-g | 28.4 ±.3 | 32.2 ±.2 | 10/0/0 | .88 | 124 ±2 | 49 ±1 | 2.51 |
| diabetes | 25.4 ±.3 | 24.9 ±.2 | 3/0/7 | 1.02 | 44.0 ±1.6 | 11.5 ±.0 | 3.81 |
| heart-c | 23.0 ±.5 | 26.8 ±.6 | 10/0/0 | .86 | 39.9 ±.4 | 20.5 ±.0 | 1.94 |
| heart-h | 21.5 ±.2 | 26.1 ±.3 | 10/0/0 | .82 | 19.1 ±.6 | 16.3 ±.3 | 1.18 |
| hepatitis | 20.4 ±.6 | 24.8 ±.3 | 10/0/0 | .82 | 17.8 ±.3 | 13.7 ±.2 | 1.30 |
| iris | 4.80 ±.17 | 4.60 ±.35 | 3/1/6 | 1.04 | 8.5 ±.0 | 12.0 ±.0 | .71 |
| labor | 19.1 ±1.0 | 15.3 ±1.6 | 3/0/7 | 1.25 | 7.0 ±.3 | 14.9 ±.1 | .47 |
| sick | 1.34 ±.03 | 2.21 ±.01 | 10/0/0 | .61 | 50.8 ±.5 | 12.0 ±.0 | 4.23 |
| sonar | 25.6 ±.7 | 28.4 ±.7 | 8/0/2 | .90 | 28.4 ±.2 | 11.1 ±.0 | 2.56 |
| vehicle | 27.1 ±.4 | 38.1 ±.3 | 10/0/0 | .71 | 135 ±2 | 16 ±0 | 8.46 |
| waveform | 27.3 ±.3 | 35.2 ±.6 | 10/0/0 | .78 | 44.6 ±.4 | 13.9 ±.0 | 3.21 |
| **average** | | | | **.91** | | | **2.44** |

It is worth noting that T2's trees are much smaller than those found by C4.5 – less than half the size, on average. This is despite the fact that tests in T2 have one more outcome (for unknown values) than the corresponding tests in C4.5.

## 6. Conclusion

The results of Section 4 show that the straightforward changes to C4.5's use of continuous attributes lead to an overall improvement in its performance on the twenty learning tasks investigated here.[5] The pruned trees are substantially smaller and somewhat more accurate, especially in the presence of irrelevant attributes. As the tasks are a representative selection from those in the UCI Repository that involve continuous attributes, similar learning tasks should also benefit. Of course, C4.5's performance on domains with continuous attributes can also be improved in other complementary ways, such as by selecting attributes (John, Kohavi, & Pfleger, 1994), exploring the space of parameter settings (Kohavi & John, 1995), or generating multiple classifiers (Breiman, 1996; Freund & Schapire, 1996).

Comparisons with a well-known global discretization scheme, and with a system that carries out a thorough search over the space of two-level decision trees, also favor the modified C4.5. However, both suggest further ways in which the system might be improved. Non-binary splits on continuous attributes make the trees easier to understand and also seem to lead to more accurate trees in some domains. It would also be interesting to investigate

---

5. The files necessary to update C4.5 Release 5 (available with Quinlan, 1993) to the new Release 8 can be obtained by anonymous ftp from ftp.cs.su.oz.au, file pub/ml/patch.tar.Z.

Kohavi's suggestion to use discretization within a tree when the local number of training cases is small.

On another tack, C4.5 has an option that affects tests on discrete attributes. Instead of the default, in which each value of the attribute is associated with a separate subtree, the values are grouped into subsets and one tree formed for each subset. Many possible subsets are explored, just as many possible thresholds for a continuous attribute are considered. The argument for the application of a penalty to tests on continuous attributes would seem to apply also to such subset tests.

## Appendix A. Description of learning tasks

| Abbrev | Domain | Cases | Classes | Attributes | |
|--------|--------|-------|---------|------|------|
| | | | | Cont | Discr |
| anneal | annealing processes | 898 | 6 | 9 | 29 |
| auto | auto insurance | 205 | 6 | 15 | 10 |
| breast-w | breast cancer (Wisc) | 699 | 2 | 9 | – |
| colic | horse colic | 368 | 2 | 10 | 12 |
| credit-a | credit screening (Aust) | 690 | 2 | 6 | 9 |
| credit-g | credit screening (Ger) | 1000 | 2 | 7 | 13 |
| diabetes | Pima diabetes | 768 | 2 | 8 | – |
| glass | glass identification | 214 | 6 | 9 | – |
| heart-c | heart disease (Clev) | 303 | 2 | 8 | 5 |
| heart-h | heart disease (Hun) | 294 | 2 | 8 | 5 |
| hepatitis | hepatitis prognosis | 155 | 2 | 6 | 13 |
| hypo | hypothyroid diagnosis | 3772 | 5 | 7 | 22 |
| iris | iris classification | 150 | 3 | 4 | – |
| labor | labor negotiations | 57 | 2 | 8 | 8 |
| letter | letter identification | 20000 | 26 | 16 | – |
| segment | image segmentation | 2310 | 7 | 19 | – |
| sick | sick euthyroid | 3772 | 2 | 7 | 22 |
| sonar | sonar classification | 208 | 2 | 60 | – |
| vehicle | silhouette recognition | 846 | 4 | 18 | – |
| waveform | waveform differentiation | 300 | 3 | 21 | – |

## Acknowledgements

This research was made possible by a grant from the Australian Research Council. The T2 system was programmed by Peter Auer and made available for these comparisons by Rob Holte. Thanks to Thierry Van de Merckt for comments on the results that led to the ablation experiments. I am also grateful for suggestions regarding the paper's content and presentation made by Ron Kohavi, Usama Fayyad, and Pat Langley. The UCI Data Repository owes its existence to David Aha and Patrick Murphy. The breast cancer data (breast-w) was provided to the Repository by Dr William H. Wolberg.